\pdfoutput=1

\documentclass[11pt]{article}

\usepackage[final]{coling}

\usepackage{times}
\usepackage{latexsym}

\usepackage[T1]{fontenc}

\usepackage[utf8]{inputenc}

\usepackage{microtype}

\usepackage{inconsolata}

\usepackage{graphicx}

\usepackage{amsmath}
\usepackage[table]{xcolor}
\usepackage{multirow}
\usepackage{enumitem}
\usepackage{titlesec}
\usepackage{booktabs}       
\usepackage{subcaption}
\usepackage{caption}

\newcommand{\RN}[1]{%
	\textup{\lowercase\expandafter{\it \romannumeral#1}}%
}

\title{DA-Cramming: Enhancing Cost-Effective Language Model Pretraining with Dependency Agreement Integration}

\author{%
\textbf{Martin Kuo},
\textbf{Jianyi Zhang}, 
\textbf{Dongting Li}, 
\textbf{Yiran Chen}\\[1ex]
\normalfont{\small Center for Computational Evolutionary Intelligence, Duke University}
}

\begin{document}
\maketitle
\begin{abstract}

Pretraining language models is still a challenge for many researchers due to its substantial computational costs. As such, there is growing interest in developing more affordable pretraining methods. One notable advancement in this area is the \textit{Cramming} technique \cite{geiping2022cramming}, which enables the pretraining of BERT-style language models using just one GPU in a single day. Building on this innovative approach, we introduce the \textit{\textbf{D}ependency \textbf{A}greement \textbf{Cramming}} (DA-Cramming), an efficient framework that integrates information about dependency agreements into the pretraining process. Unlike existing methods that leverage similar semantic information during \textit{finetuning}, our approach represents a pioneering effort focusing on enhancing the foundational language understanding with semantic information during \textit{pretraining}. We meticulously design a dual-stage pretraining workflow with four dedicated submodels to capture representative dependency agreements at the chunk level, effectively transforming these agreements into embeddings to benefit the pretraining. Extensive empirical results demonstrate that our method significantly outperforms previous methods across various tasks. 

\end{abstract}

\section{Introduction}
\label{sec:intro}


Pretrained language models, such as BERT \cite{devlin2018bert}, RoBERTa \cite{liu2019roberta}, ALBERT \cite{lan2019albert}, and GPT series \cite{radford2018improving, radford2019language, brown2020language}, have recently become prominent across various natural language processing (NLP) tasks. However, the pretraining of these models typically requires substantial computational resources, making them inaccessible to many researchers and practitioners \cite{geiping2022cramming}. To address this challenge, there has been growing interest in developing more afforable pretraining methods that can operate under limited GPU resources. A notable example is the \textit{Cramming} technique \cite{geiping2022cramming}, which explores the feasibility of pretraining BERT on a single GPU within a day. By refining model components—such as eliminating biases from linear layers and adopting scaled sinusoidal positional embeddings—this approach achieves efficient pretraining without sacrificing model size.

Although the \textit{Cramming} technique approaches the performance levels of standard pretraining on BERT, a performance gap still remains. To further narrow this gap and enhance the interpretability of pretraining, we incorporate semantic information, the dependency agreement, into the pretraining to capture linguistic relationships more effectively. Notably, while previous efforts have introduced similar syntactic or semantic information during the fine-tuning phase \cite{bai2021syntax, zanzotto2020kermit}, they neglect the critical pretraining phase, where foundational language understanding is developed. For instance, \cite{bai2021syntax} and \cite{zanzotto2020kermit} incorporate syntax trees into fine-tuning a pretrained BERT model. Another study by \cite{vig2019multiscale} enables the visualization of attention information to understand which words are semantically significant for the model's behaviors.

\begin{figure*}[!ht]
    \centering
    \begin{minipage}{0.47\textwidth} 
        \centering
        \includegraphics[width=0.95\linewidth]{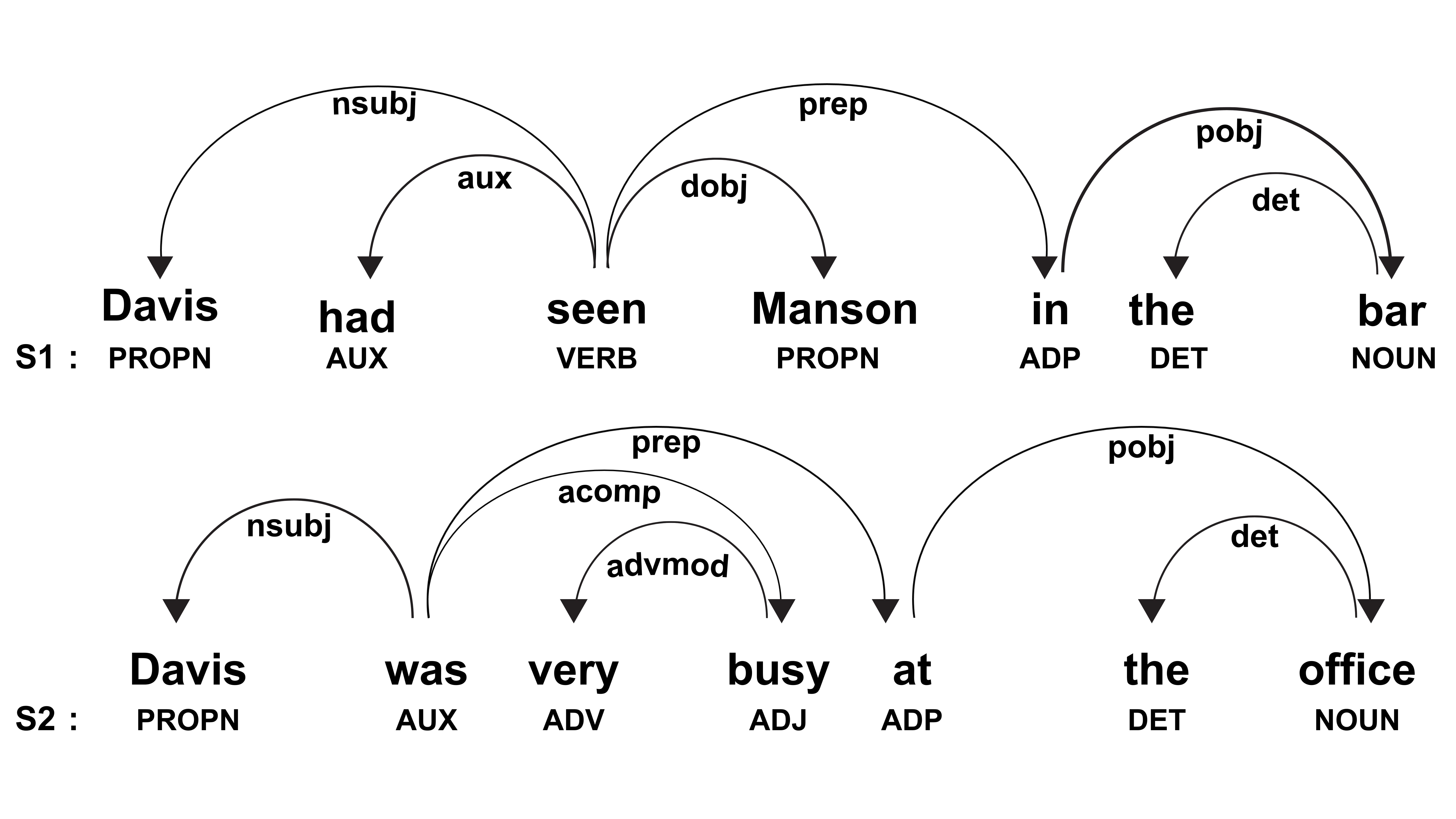} 
        \captionof{figure}{Examples of dependency structures of two sentences (Left: Example 1. Right: Example 2).} 
        \label{dependency_example_1}
        \vspace{-0.1cm}
    \end{minipage}%
    \hfill 
    \begin{minipage}{0.47\textwidth} 
        \centering
\resizebox{\linewidth}{!}{
\begin{tabular}{|l|l|l|l|l|}
\hline
\multicolumn{1}{|c|}{\begin{tabular}[c]{@{}c@{}}\textbf{Dependency}\\ \textbf{Agreement}\end{tabular}}  & \multicolumn{1}{|c|}{\textbf{Grammar}} & \multicolumn{1}{|c|}{\textbf{Description}}               & \multicolumn{1}{|c|}{\textbf{Example 1}}                       & \multicolumn{1}{|c|}{\textbf{Example 2}}                         \\ \hline
\multicolumn{1}{|c|}{\multirow{4}{*}{\textbf{SV}}}   & \multicolumn{1}{|c|}{\textbf{nsubj}}            & \multicolumn{1}{|c|}{\textbf{Nominal subject}}           & \multicolumn{1}{|c|}{\multirow{4}{*}{\textbf{Davis had seen}}} & \multicolumn{1}{|c|}{\multirow{4}{*}{\textbf{Davis was}}}        \\ \cline{2-3}
                                           & \multicolumn{1}{|c|}{\textbf{nsubjpass}}        & \multicolumn{1}{|c|}{\textbf{Nominal subject (passive)}} &                                 &                                   \\ \cline{2-3}
                                           & \multicolumn{1}{|c|}{\textbf{csubj}}            & \multicolumn{1}{|c|}{\textbf{Clausal subject}}           &                                 &                                   \\ \cline{2-3}
                                           & \multicolumn{1}{|c|}{\textbf{csubjpass}}        & \multicolumn{1}{|c|}{\textbf{Clausal subject (passive)}} &                                 &                                   \\ \hline
\multicolumn{1}{|c|}{\textbf{DOBJ}}                  & \multicolumn{1}{|c|}{\textbf{dobj}}             & \multicolumn{1}{|c|}{\textbf{Direct Object}}             & \multicolumn{1}{|c|}{\textbf{seen Mason}}                      &                                   \\ \hline
\multicolumn{1}{|c|}{\textbf{POBJ}}                  & \multicolumn{1}{|c|}{\textbf{pobj}}             & \multicolumn{1}{|c|}{\textbf{Object of preposition}}     & \multicolumn{1}{|c|}{\textbf{in the bar}}                      & \multicolumn{1}{|c|}{\textbf{at the office}}                       \\ \hline
\multicolumn{1}{|c|}{\multirow{5}{*}{\textbf{COMP}}} & \multicolumn{1}{|c|}{\textbf{acomp}}            & \multicolumn{1}{|c|}{\textbf{Adjectival complement}}     & \multicolumn{1}{|c|}{\multirow{5}{*}{}}               & \multicolumn{1}{|c|}{\multirow{5}{*}{\textbf{was very busy}}} \\ \cline{2-3}
                                           & \multicolumn{1}{|c|}{\textbf{xcomp}}            & \multicolumn{1}{|c|}{\textbf{Open clausal complement}}   &                                 &                                   \\ \cline{2-3}
                                           & \multicolumn{1}{|c|}{\textbf{ccomp}}            & \multicolumn{1}{|c|}{\textbf{Clausal complement}}        &                                 &                                   \\ \cline{2-3}
                                           & \multicolumn{1}{|c|}{\textbf{pcomp}}            & \multicolumn{1}{|c|}{\textbf{Complement of preposition}} &                                 &                                   \\ \cline{2-3}
                                           & \multicolumn{1}{|c|}{\textbf{attr}}             & \multicolumn{1}{|c|}{\textbf{Attribute}}                 &                                 &                                   \\ \hline
\end{tabular}}
        \captionof{table}{\label{dependency_agreement_group}
The rule for grouping Dependency Agreement (SV, DOBJ, POBJ, COMP) is based on dependency grammar and the Lexical Approach.} 
    \end{minipage}
    \vspace{-0.3cm}
\end{figure*}


In contrast, our framework, termed {\textit{\textbf{D}ependency \textbf{A}greement \textbf{Cramming}}} (DA-Cramming), integrates the syntactic and semantic insights from the start of pretraining. Inspired by the Lexical Approach \cite{lewis1993lexical} and dependency grammar \cite{de2019dependency}, our dual-stage framework emphasizes language learning in chunks—such as subject-verb and verb-object relations—rather than individual words, thereby enhancing fluency and idiomatic expression. In the first stage, we train submodels to generate embeddings that encapsulate these syntactic and semantic relationships, which subsequently benefits the language model during the second stage of pretraining.

Our contribution can be summarized as follows,
\begin{itemize}
\item We propose the first framework that weaves syntactic information into the pretraining process, aligning with theories such as the Lexical Approach and dependency grammar.\vspace{-0.1cm}
\item Our approach significantly boosts the performance of Cramming on downstream tasks, yielding an increase of 1.06\% for the average GLUE score, without requiring additional computational resources on GPUs. \vspace{-0.1cm}
\item Our framework improves the interpretability of pretrained language models, especially for natural language understanding tasks. \vspace{-0.1cm}
\end{itemize}

\section{Dependency Agreement Cramming} \label{method}
\subsection{Dependency Agreement Rule} \label{Detailed_Agreement_Rule}

Dependency grammar \cite{de2019dependency} is a grammatical structure based on the concept of dependency relations. We use the spacy dependency parser \cite{spacy2} based on the rules presented in \cite{honnibal2015improved} and \cite{nivre-nilsson-2005-pseudo} to convert sentences into dependency structure, as illustrated in Figure~\ref{dependency_example_1}. Then we determine the dependency agreement, based on the definition of agreement \cite{Cysouw+2011+153+160}, for each word following our defined grouping rules in Table~\ref{dependency_agreement_group}. We define four distinct types of dependency agreements: subject/verb (SV) agreement, verb/direct object (DOBJ) agreement, verb/object of preposition agreement (POBJ), and complement agreement (COMP).

The grouping operation is grounded in the Lexical Approach, as proposed by Lewis \cite{lewis1993lexical}, which suggests that language consists of 'chunks' that enable the production of continuous and coherent text. This concept has been substantiated by subsequent research, such as that by Schmitt \cite{schmitt2000keyconceptsinelt}, which demonstrates that it is more cognitively efficient for our brains to recall language information in chunks rather than as individual words. Furthermore, Scott \cite{scott2019cambridge} has found that learning languages in chunks benefits the language acquisition.

\subsection{Stage 1: Pretrain the Dependency Agreement Submodels} \label{First_Stage}
\begin{figure}[!ht]
        \centering
        \includegraphics[width=\linewidth]{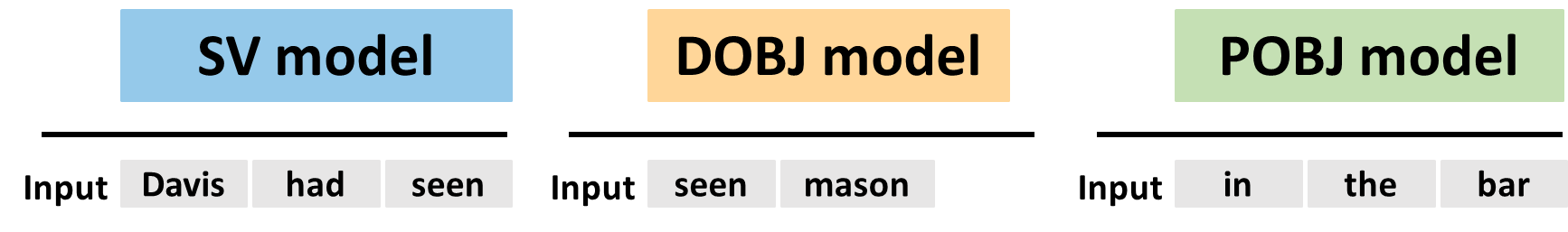} 
        \captionof{figure}{The Stage 1 of DA-cramming. Train each dependency agreement submodel using its corresponding agreement dataset. 
        }
        \label{agreement_submodel}
        \vspace{-0.1cm}
\end{figure}
In the initial stage, we segment each sentence for pretraining into four distinct agreement types: SV, DOBJ, POBJ, and COMP, creating corresponding sub-datasets for each type. Each agreement submodel, tailored to its specific dataset, is trained using a two-layer architecture based on the Crammed Bert \cite{geiping2022cramming}. The inputs to these submodels are words categorized by predefined agreement rules.  The training employs a “masked language model” (MLM) pre-training objective \cite{devlin2018bert} for each submodels.

\subsection{Stage 2: Combine Dependency Agreement Submodels}
\label{Second_Stage}
\vspace{-0.2cm}
\begin{figure}[!ht]
        \centering
        \includegraphics[width=\linewidth]{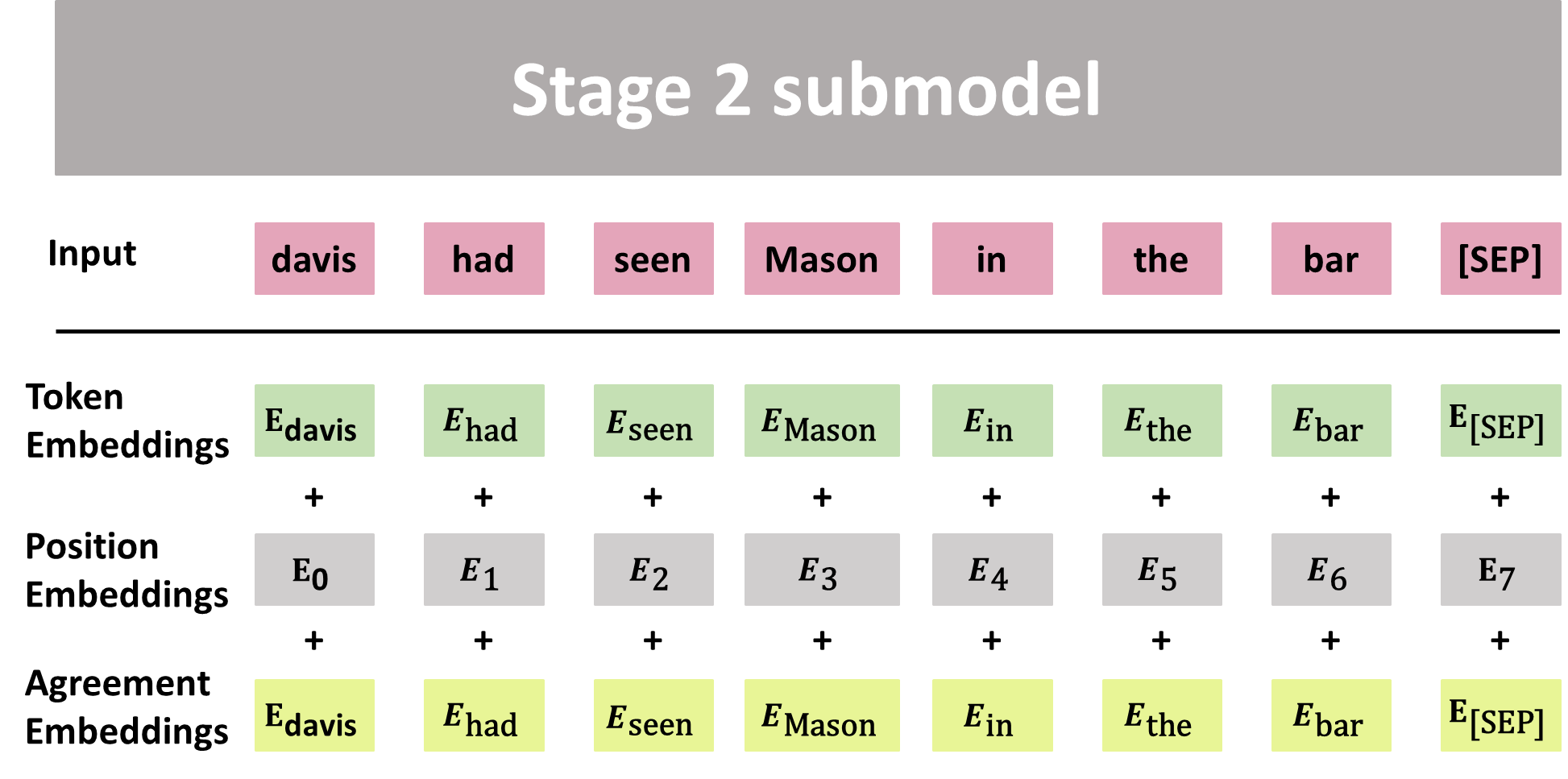} 
        \captionof{figure}{Stage 2 model is based on Crammed BERT architecture} 
    \label{pretrain_embedding}
    \vspace{-0.2cm}
\end{figure}
The second stage utilizes the four trained agreement submodels to generate robust agreement embeddings. After applying agreement rules to a tokenized input sequence {\( T=(T_0, T_1, ..., T_{L-1}) \)}, where \( L \) denotes the input's maximum length, each word is processed through its corresponding submodel. The outputs are then weighted by their respective agreement attention scores and aggregated to form the agreement embedding, which is normalized using Layer Norms \cite{baevski2018adaptive, xiong2020layer}: $Agr(T)=$
{\begin{align}
LayerNorm(Attn(SV(T))+Attn(DOBJ(T)) \nonumber \\+Attn(POBJ(T))+Attn(COMP(T))) \nonumber.
\vspace{-0.8cm}
\end{align}}
Then, this agreement embedding is integrated with token and positional embeddings to construct the composite input embedding for the second stage model, which has the same architecture as Crammed BERT \cite{geiping2022cramming}:
{\begin{align}
Embed(T) = Token(T) + Pos(T) + Agr(T),\nonumber
\vspace{-0.4cm}
\end{align}} where $Token$, $Pos$,  and $Agr$ correspond to the Token embedding, Positional embedding, Agreement embedding. Then, we train the model with the MLM objective.

\section{Experiments}
\subsection{Implementation Details}
\paragraph{Dataset} We use the Bookcorpus \cite{Zhu_2015_ICCV} and Wikipedia (20220301.en) datasets \cite{wikidump} for the pretraining of our framework. Firstly, we apply the agreement rule (as described in Section ~\ref{Detailed_Agreement_Rule}) to these datasets, classifying words into four distinct subdatasets (i.e. SV, DOBJ, POBJ, COMP). We utilize these four subdatasets to train each submodel during our framework's first stage pretraining. Then we adopt the original datasets for the stage 2 pretraining. We use GLUE \cite{wang2018glue} as our finetuning dataset and evaluation benchmark, which includes several natural language understanding tasks.

\paragraph{DA-Cramming \& Baselines}
In the first stage pretraining, each submodel follows a 2-layer architecture based on Crammed BERT \cite{geiping2022cramming} and applies the FLASH mechanism \cite{dao2022flashattention}. In our framework's second-stage pretraining, we combine these four 2-layer submodels, followed by the Stage 2 submodel which has 14 layers of architecture based on Crammed BERT \cite{geiping2022cramming} and utilizing the FLASH mechanism, resulting in a whole model of 16 layers, same as Crammed BERT. More pretraining details are in the appendix Section ~\ref{first_stage_pretraining_details}.

We adopt the Crammed BERT \cite{geiping2022cramming}, it's pretraining details is in the appendix section ~\ref{crammed_bert_pretraining_details}, as the main baseline to verify our framework which includes the syntactic information into the pretraining process can improve the performance of pretraining. Moreover, to compare with other approaches which incorporate syntax information into the fine-tuning, we implement Syntax Crammed BERT~\cite{bai2021syntax} and Crammed
BERT + KERMIT~\cite{zanzotto2020kermit} as other representative baselines, which incorporate the syntax tree information into the finetuning of Crammed BERT.

\begin{table*}[t!]
\centering
\renewcommand{\arraystretch}{1.5}  
\resizebox{0.95\textwidth}{!}{
\begin{tabular}{l|ccccccccc}
\hline
\multicolumn{1}{c|}{}  & RTE                    & CoLA                   & MRPC                   & MNLI(m/mm)                             & QQP                    & SST-2                 & STSB                   & QNLI                  & GLUE                   \\ \hline
Crammed BERT           & 54.39 ± 1.269          & 41.53 ± 0.191          & 84.70 ± 0.070          & 81.85 ± 0.061 / 82.21 ± 0.447          & 86.78 ± 0.101          & 91.59 ± 1.172         & 84.13 ± 0.615          & \textbf{89.48 ± 0.13} & 77.41 ± 0.223          \\
Syntax Crammed BERT    & 54.63 ± 1.460          & 15.01 ± 2.376          & 81.62 ± 0.357          & 79.09 ± 0.585 / 79.24 ± 0.005          & 85.67 ± 0.159          & 88.61 ± 0.827         & 81.44 ± 0.891          & 84.62 ± 0.503         & 72.21 ± 0.192          \\
Crammed BERT + KERMIT & 53.43±0.360            & 41.72±0.852            & 84.90±0.541            & 80.98 $\pm$ 0.062 / 81.31 $\pm$ 0.096   & 86.74±0.015            & 91.97 ± 1.65          & 82.91    ± 0.325       & 85.89 $\pm$ 0.267     & 76.65 $\pm$ 0.157      \\
DA-Crammed BERT (ours) & \textbf{57.52 ± 1.362} & \textbf{42.48 ± 1.145} & \textbf{86.96 ± 0.440} & \textbf{81.87 ± 0.119 / 82.24 ± 0.266} & \textbf{86.93 ± 0.102} & \textbf{91.74 ± 0.46} & \textbf{85.10 ± 0.521} & 89.32 ± 0.170         & \textbf{78.24 ± 0.109} \\ \hline
Crammed BERT + SMART  & 55.72±0.755            & 41.2±1.091             & 86.94±0.382            & 82.82±0.046/82.88±0.083                & 87.61±0.171            & 92.51±0.173           & 85.25±0.327            & 90.16±0.14            & 78.34±0.209            \\
DA-Crammed BERT + SMART (ours) & 58.12±0.99             & 44.23±1.23             & 87.36±0.375            & 82±0.045/82.92±0.083                   & 87.73±0.159            & 92.52±0.19            & 85.42±0.37             & 89.97±0.121           & 78.92±0.189            \\ \hline
\end{tabular}}
\caption{\label{glue_exp}
GLUE-dev performance.
}
\end{table*}

\subsection{Evaluating Finetuning on GLUE}
We evaluate Crammed BERT, Syntax Crammed BERT, Crammed BERT + KERMIT, and our DA-Crammed BERT on the GLUE dataset using three different random seeds. The detailed finetuning settings are in the appendix Section ~\ref{finetuning_settings}. The average evaluation results are presented in Table~\ref{glue_exp}.
The results demonstrate that our framework exhibits superior performance in specific tasks. For example, in the RTE task, our framework surpasses Crammed BERT by 3.13\%, and in the MRPC task, it outperforms Crammed BERT by 2.26\%. Furthermore, our framework outperforms Crammed BERT in 7 out of 8 downstream tasks and achieves an average GLUE score improvement of 0.83\%.
Contrastingly, the underperformance of Syntax Crammed BERT could be attributed to machanism of Crammed BERT, which potentially obstructs the efficient deployment of syntax trees during fine-tuning. Besides, our DA-Crammed BERT also outperforms Crammed BERT + KERMIT.

\begin{figure*}[ht]
    \centering
    \begin{subfigure}{0.48\textwidth}
        \includegraphics[width=\linewidth]{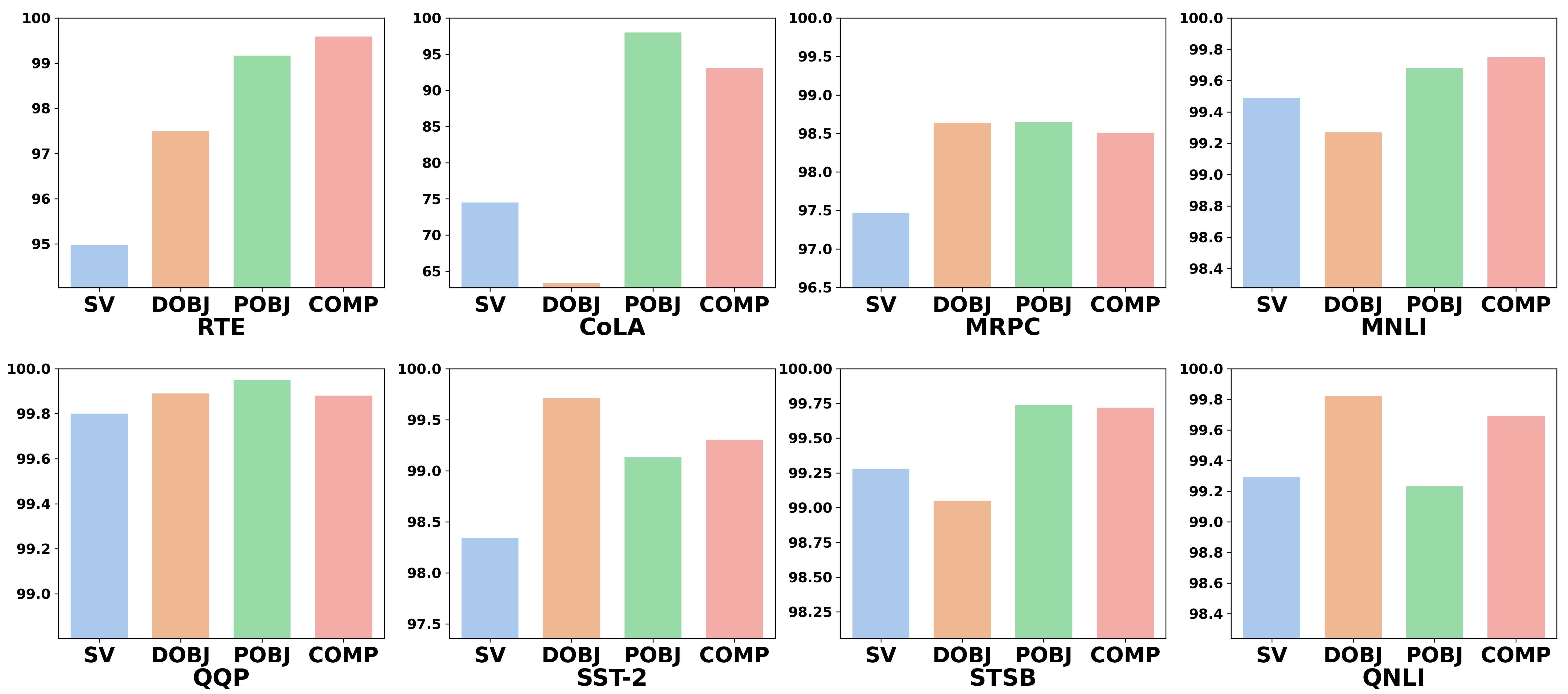}
        \caption{Submodel Significance Analysis}
        \label{agreement_ablation}
    \end{subfigure}
    \hfill 
    \begin{subfigure}{0.48\textwidth}
        \includegraphics[width=\linewidth]{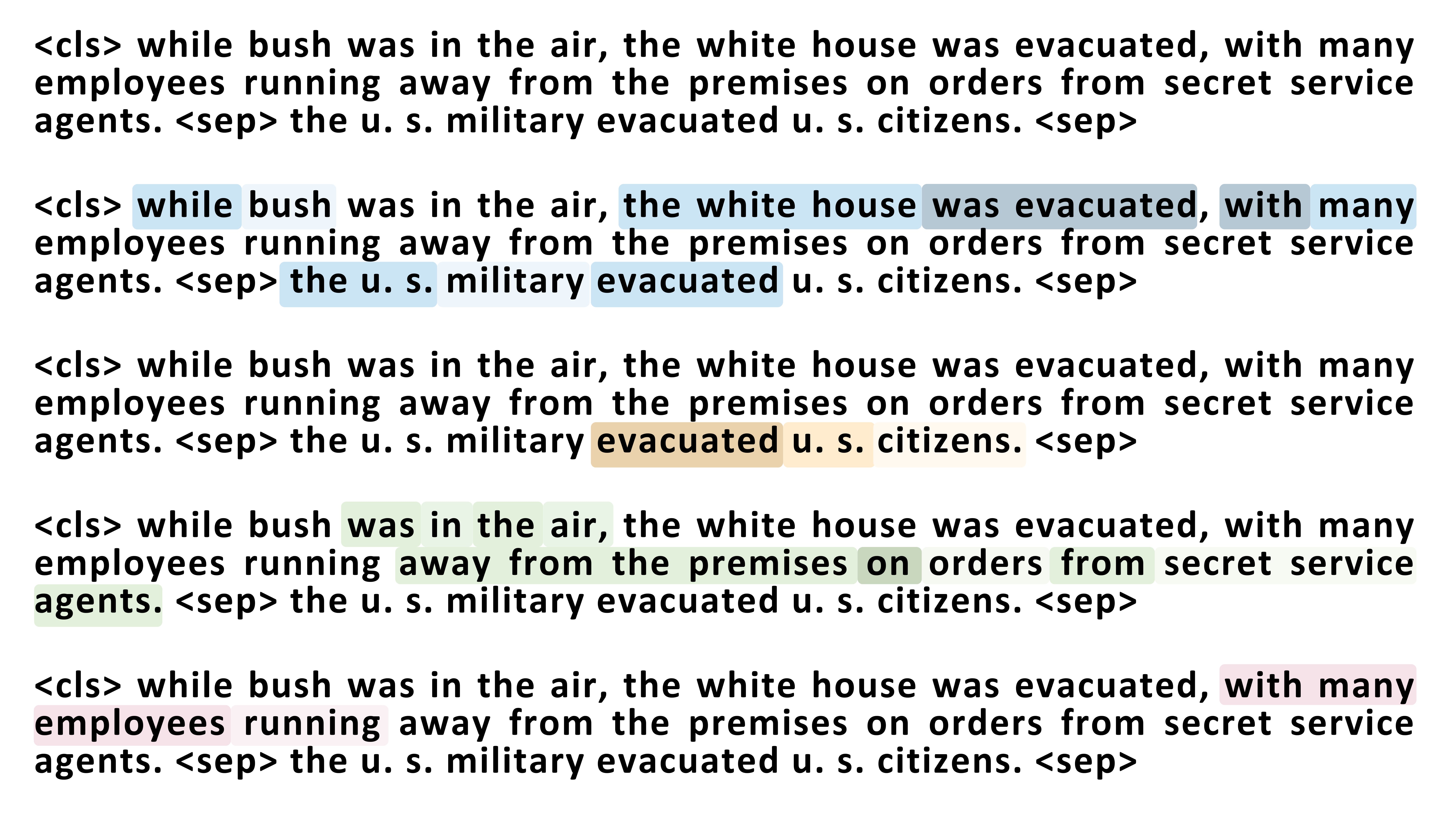}
        \caption{Submodel Attention Analysis}
        \label{sentence_agreement_ablation}
    \end{subfigure}
    \caption{Aditional Analysis}
    \label{fig:combined_ablation}
    \vspace{-0.5cm}
\end{figure*}



\subsection{Ablation Studies and Additional Analysis}



\subsubsection{Evaluating Advanced Finetuning Method}

We aim to investigate whether applying a fine-tuning method, which lacks a relationship with syntactic knowledge, can improve the performance of our method further on downstream tasks. Therefore, we implemented the SMART \cite{jiang2019smart} — a more advanced finetuning technique, using the GLUE dataset and initializing it with three distinct random seeds. The aggregated results consistently indicate that with the finetuning method SMART, our model still outperforms Crammed BERT , as shown in Table~\ref{glue_exp}.

\subsubsection{Submodel Significance Analysis}




To evaluate the impact of each agreement submodel, we conducted an ablation study by removing individual submodels from various GLUE tasks and observed the accuracy drop for each task as shown in Figure~\ref{agreement_ablation}. A larger drop indicates a higher importance of the submodel for the task.

In Figure~\ref{agreement_ablation}, the task performance is at 100\% with all four submodels. Removing a submodel results in a performance drop to (100-x)\%, reflecting the remaining accuracy. This figure highlights the critical role of the SV submodel across all tasks and underscores the importance of both SV and DOBJ submodels in the RTE task, which demands understanding of "Who" and "What" actions \cite{hart2002five}. The SV identifies the subject of an action, and the DOBJ identifies the action's object, underscoring their pivotal roles in RTE.

\subsubsection{Submodel Attention Analysis}
We investigate the interpretation from chunks perspective with a sentence from an RTE task: "While Bush was in the air, the White House was evacuated, with many employees running away from the premises on orders from secret service agents. The U.S. military evacuated U.S. citizens." We analyze this scenario across four syntactic SV, DOBJ,POBJ, and COMP—as depicted in Figure~\ref{sentence_agreement_ablation}. In this figure, the attention scores are highlighted, with darker colors indicating higher focus. Our analysis reveals that the verb "evacuated" and its agent are central to the entailment analysis in both the SV and DOBJ models, emphasizing the action and the primary actors. In contrast, the POBJ model assigns moderate attention to the phrase "from the premises," pinpointing its lesser yet notable role. The COMP model, however, allocates minimal attention to ancillary elements such as "with many employees running," underscoring their negligible impact on the entailment outcome. This syntactic ordering (SV, DOBJ, POBJ, then COMP) correlates with the significance levels outlined in Figure~\ref{agreement_ablation}, illustrating the importance in syntactic analysis for entailment tasks, where SV and DOBJ are prioritized over POBJ and COMP.
\vspace{-0.1cm}

\section{Conclusion}
\vspace{-0.2cm}
We introduce Dependency Agreement Cramming, a dual-stage pertaining framework that aligns with principles of the Lexical Approach and dependency grammar. As evident from our evaluation on the GLUE benchmark, DA-Cramming shows considerable enhancements across several tasks without requiring additional computational resource. Further examinations elucidate the pivotal role our approach plays in heightening the interpretability for natural language understanding tasks. As for future work, we aim to scale our framework to accommodate larger language models such as GPTs, while also exploring its potential with more datasets.

\section{Limitations}
In our work, we have identified concerns regarding users' data privacy. Currently, our framework involves inputting users' actual words in the second stage, which poses a risk of the cloud server gaining access to users' actual words and potentially compromising their privacy. To address this issue in our future work, we intend to implement a clear separation between first stage and second stage inputs. This means that users' actual words will only be inputted in the first stage, eliminating the need to upload them to the cloud.

\bibliography{custom}

\appendix

\section{Experiments}

\subsection{Implementation Details}

\paragraph{First stage pretraining details}
\label{first_stage_pretraining_details}

We set the micro-batch size to 128 and accumulate it into larger batch sizes of 4096. Training for each submodel is performed using a triangular learning rate scheduler \cite{geiping2022cramming}, commencing with a learning rate of 25e-5 and training for one epoch. Additionally, we employ the WordPiece tokenizer with a vocabulary size of 32768 \cite{wu2016google}.

Additionally, each agreement's submodel has a different maximum input length, as specified in Table~\ref{agreement_model_input_length}. If the sequence length exceeds the maximum input length, we truncate the sequence accordingly.

\begin{table}[ht]
\centering
\resizebox{\linewidth}{!}{%
\begin{tabular}{lllll}
\hline
& \textbf{SV} & \textbf{DOBJ} & \textbf{POBJ} & \textbf{COMP} \\ \hline
\textbf{Dataset Size} & 35M & 35M & 35M & 35M \\
\textbf{Max Input Length} & 19 & 21 & 10 & 23 \\
\textbf{Length Percentile} & 98\% & 96\% & 98\% & 97\% \\ \hline
\end{tabular}%
}
\vspace{2mm}
\caption{\label{agreement_model_input_length}
This table summarizes the attributes of the Stage 1 pretraining dataset, including Dataset Size, Max Input Length and its corresponding Length Percentile. Agreement model input length: we analyze the length of each agreement in the BookCorpus-Wikipedia dataset \cite{Zhu_2015_ICCV} \cite{wikidump} and configure the maximum input length for each agreement to accommodate over 95\% of the chunks, ensuring they fall within this specified length.
}
\end{table}

\begin{figure*}[t!]
\includegraphics[width=0.9\textwidth,height=4.5cm]{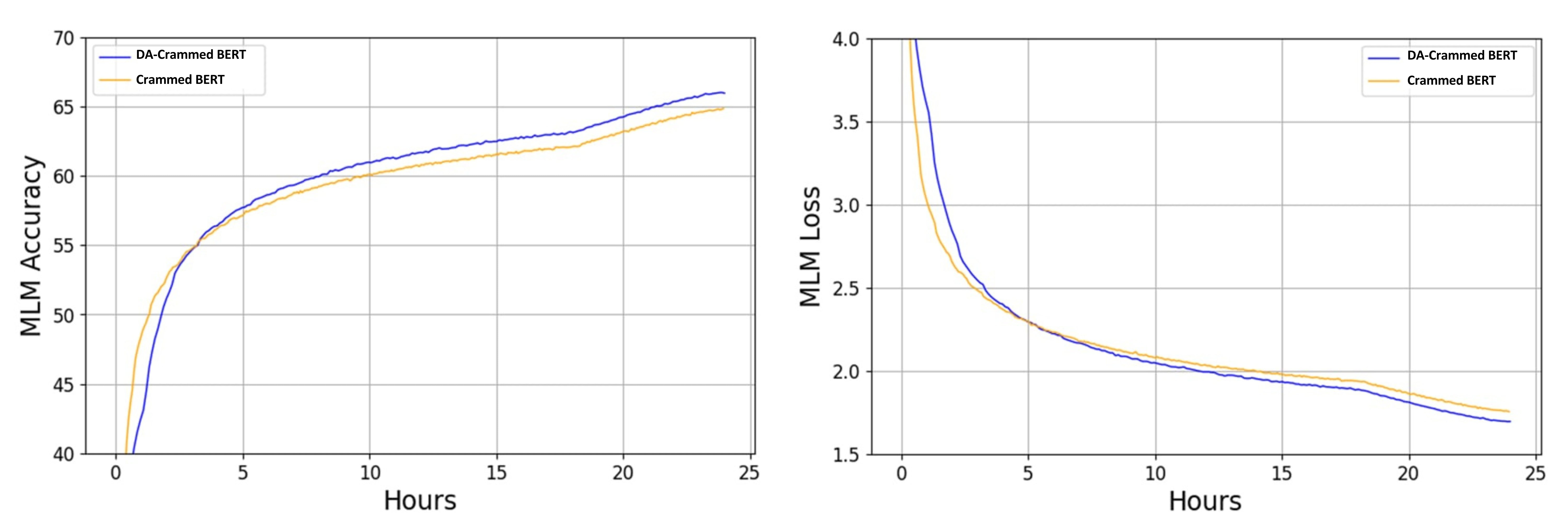}
  \caption{Summary of Pretraining Stage 2. Left: MLM Accuracy. Right: MLM loss}
  \label{accuracy_loss_pic}
\end{figure*}
\begin{figure*}[ht]
  \includegraphics[width=\linewidth]{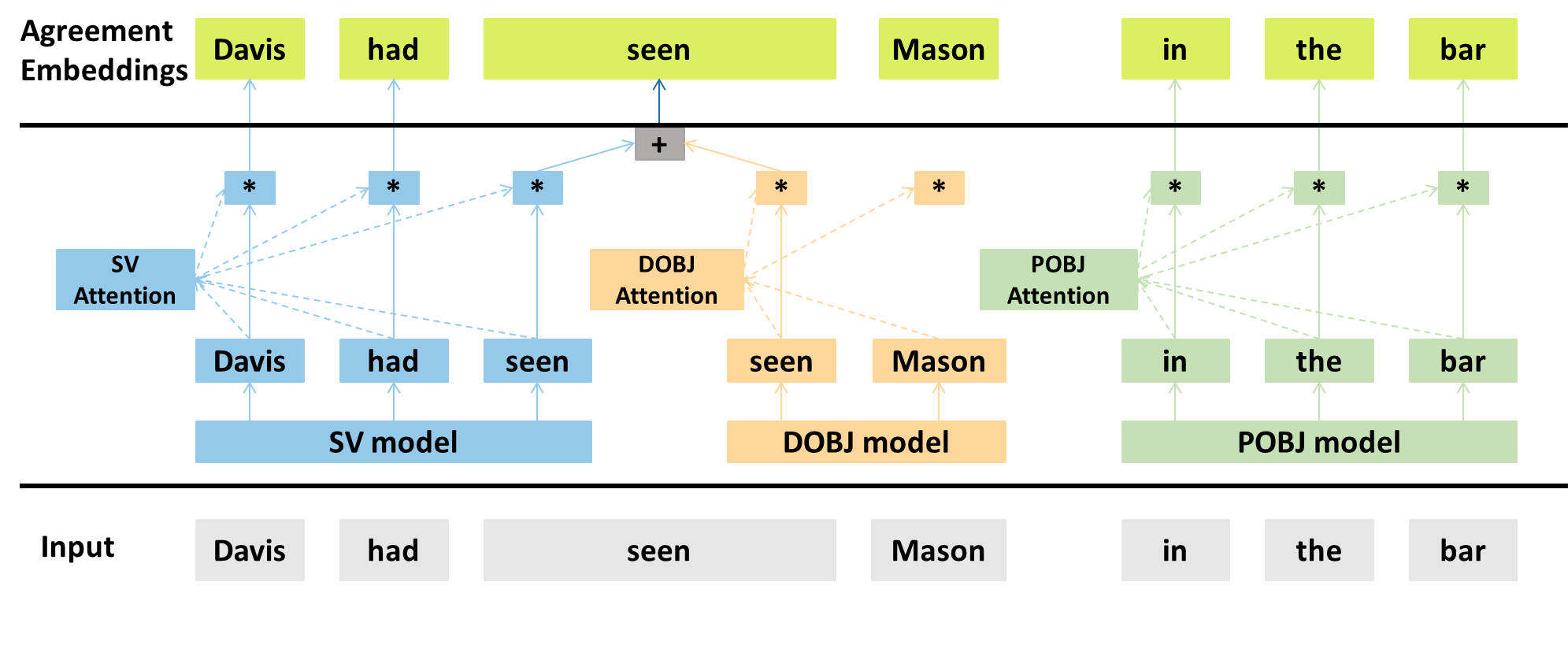}
  \caption{Dependency Agreement embedding generation in Stage 2: We combine four agreement submodels and applied an attention mechanism to each agreement.}
  \label{agreement_embedding_generation}
\end{figure*}
\paragraph{Second stage pretraining details}
\label{second_stage_pretraining_details}
We adopt the same batch size as Stage 1. The training of the model employs a triangular learning rate scheduler \cite{geiping2022cramming}, starting with a learning rate of 25e-5 and completing the training within 24 hours. 

\paragraph{Crammed BERT pretraining details}
\label{crammed_bert_pretraining_details}
We configure it with 16 layers and employ the FLASH mechanism \cite{dao2022flashattention}. We utilize the same triangular learning rate scheduler and batch size following the paper \cite{geiping2022cramming}. 

\paragraph{Finetuning settings} \label{finetuning_settings} During the finetuning phase, we perform finetuning on all GLUE datasets for 5 epochs with 16 as batch size, employing a cosine decay learning rate scheduler. The initial learning rates are determined through a grid search spanning from 1e-4 to 9e-5.

\section{Details of Stage 1 \& Stage 2 Pretraining}

\begin{table}[ht]
\centering
\resizebox{\linewidth}{!}{%
\begin{tabular}{lllll}
\toprule
& \textbf{SV} & \textbf{DOBJ} & \textbf{POBJ} & \textbf{COMP} \\
\midrule
\textbf{Acc (\%)} & 38 & 36 & 43 & 43 \\
\bottomrule
\end{tabular}
}
\caption{Summary of Pretraining Stage 1. This table shows the accuracy percentages for different submodels in the initial stage of pretraining.}
\label{stage1_dataset}
\end{table}

In the first stage of our methodology, we compile datasets for each agreement category with 35 million chunks. The MLM accuracy for these datasets is summarized in Table \ref{stage1_dataset}. As indicated, the MLM accuracy figures are not particularly high, an expected outcome due to the increased word combinations inherent in the chunk-level analysis compared to sentence-level. To illustrate, consider a Subject-Verb (SV) agreement where 'Tom' is the subject, and the verb is masked. In such a case, both 'Tom eats' and 'Tom runs' would be plausible completions. Besides, subsequent results from Stage 2 pretraining and fine-tuning indicate that an MLM accuracy of around 40\% at Stage 1 is sufficient.

Figure \ref{accuracy_loss_pic} presents the MLM accuracy and loss in Stage 2. Evidently, DA-Crammed BERT outperforms Crammed BERT on both measures, endorsing the efficacy of our Stage 1 pretraining and the four agreement submodels.

\section{Agreement Embedding randomization Case Study}

We aimed to assess the efficacy of agreement embedding within DA-Crammed BERT. To this end, we carried out an experiment employing random vectors for agreement embedding. The findings reveal that the modules developed during our initial stage surpass those equipped with random embeddings in performance, as showed in Table~\ref{random_glue_exp}.

\begin{table}[ht]
\centering
\renewcommand{\arraystretch}{1.5}  
\resizebox{\linewidth}{!}{
\begin{tabular}{l|cc}
\hline
\multicolumn{1}{c|}{Metric} & DA-Crammed BERT (ours) & DA-Crammed BERT (random) \\ \hline
RTE           & \textbf{57.52 ± 1.362}  & 53 ± 1.5         \\
CoLA          & \textbf{42.48 ± 1.145}  & 0.0 ± 0.0        \\
MRPC          & \textbf{86.96 ± 0.440}  & 82.1 ± 0.3       \\
MNLI(m/mm)    & \textbf{81.87 ± 0.119 / 82.24 ± 0.266}  & 51.1 ± 14 / 51.5 ± 14.3 \\
QQP           & \textbf{86.93 ± 0.102}  & 77.7 ± 6.5       \\
SST-2         & \textbf{91.74 ± 0.46}   & 82.8 ± 1.9       \\
STSB          & \textbf{85.10 ± 0.521}  & 11.9 ± 1.4       \\
QNLI          & \textbf{89.32 ± 0.170}           & 67.2 ± 11.2      \\
GLUE          & \textbf{78.24 ± 0.109}  & 53.1 ± 4.4       \\ \hline
\end{tabular}}
\caption{\label{random_glue_exp}
Comparison in GLUE-dev performance between DA-Crammed BERT (ours) and DA-Crammed BERT (random)
}
\end{table}

\section{Agreement Embedding construction}
Agreement embedding is consists of four submodels and combined with each corresponding Attention mechanisms, as shown in Figure~\ref{agreement_embedding_generation}.

\end{document}